\documentclass[11pt, letterpaper]{template}
\usepackage[authoryear, round]{natbib}
\usepackage{hyperref}[citecolor=magenta]
\usepackage{graphicx}

\hypersetup{
    colorlinks=true,
    citecolor=blue,
    linkcolor=blue,
    urlcolor=blue
}

\definecolor{darkblue}{rgb}{0, 0, 0.5}
\usepackage{url}            % simple URL typesetting
\usepackage{booktabs}       % professional-quality tables
\usepackage{amsfonts}       % blackboard math symbols
\usepackage{nicefrac}       % compact symbols for 1/2, etc.
\usepackage{microtype}      % microtypography
\usepackage{xcolor}         % colors

\usepackage{lineno}

\usepackage{enumitem}       % [leftmargin=20pt]
\usepackage{subfig}         % subfloat
\usepackage{multirow}       % 
\usepackage{xspace}         % xspace
\usepackage{wrapfig}        % wrapfigure
\usepackage{makecell}       % divide row in table
\usepackage{colortbl}
\usepackage{tcolorbox}      % tcolorbox
\tcbuselibrary{breakable}
\usepackage{bbm}            % indicator function (\mathbbm)
\usepackage{adjustbox}

\usepackage{pifont}         % \ding{51}, \ding{55}
\usepackage{bbding}         % \Checkmark, \XSolidBrush

\newcommand{\ourmethod}{GenPRM\xspace}

\newcommand{\mycolor}{cyan!10}

\usepackage{circledsteps}

\newcommand\blfootnote[1]{%
  \begingroup
  \renewcommand\thefootnote{}\footnote{#1}%
  \addtocounter{footnote}{-1}%
  \endgroup
}

\title{GenPRM: Scaling Test-Time Compute of Process Reward Models via Generative Reasoning}

\author[1,3$*$]{Jian Zhao}
\author[1,2$*\dag$]{Runze Liu}
\author[1]{Kaiyan Zhang}
\author[3]{Zhimu Zhou}
\author[4]{Junqi Gao}
\author[4]{Dong Li}
\author[1]{Jiafei Lyu}
\author[4]{Zhouyi Qian}
\author[2$\ddag$]{Biqing Qi}
\author[1$\ddag$]{Xiu Li}
\author[1,2$\ddag$]{Bowen Zhou}

\affil[1]{Tsinghua University}
\affil[2]{Shanghai AI Laboratory}
\affil[3]{BUPT}
\affil[4]{Harbin Institute of Technology}
\begin{abstract}
Recent advancements in Large Language Models (LLMs) have shown that it is promising to utilize Process Reward Models (PRMs) as verifiers to enhance the performance of LLMs. However, current PRMs face three key challenges: (1) limited process supervision and generalization capabilities, (2) dependence on scalar value prediction without leveraging the generative abilities of LLMs, and (3) inability to scale the test-time compute of PRMs. In this work, we introduce GenPRM, a generative process reward model that performs explicit Chain-of-Thought (CoT) reasoning with code verification before providing judgment for each reasoning step. To obtain high-quality process supervision labels and rationale data, we propose Relative Progress Estimation (RPE) and a rationale synthesis framework that incorporates code verification. Experimental results on ProcessBench and several mathematical reasoning tasks show that GenPRM significantly outperforms prior PRMs with only \textbf{23K} training data from \textbf{MATH} dataset. Through test-time scaling, a \textbf{1.5B} GenPRM outperforms \textbf{GPT-4o}, and a \textbf{7B} GenPRM surpasses \textbf{Qwen2.5-Math-PRM-72B} on ProcessBench. Additionally, GenPRM demonstrates strong abilities to serve as a critic model for policy model refinement. This work establishes a new paradigm for process supervision that bridges the gap between PRMs and critic models in LLMs. Our code, model, and data are available in \url{https://ryanliu112.github.io/GenPRM}.
\end{abstract}

\begin{document}

\blfootnote{$^*$ Equal contribution}
\blfootnote{$^\dag$ Project lead \& Work done during an internship at Shanghai AI Laboratory}
\blfootnote{$^\ddag$ Corresponding authors: Biqing Qi (qibiqing@pjlab.org.cn), Xiu Li (li.xiu@sz.tsinghua.edu.cn), and Bowen Zhou (zhoubowen@tsinghua.edu.cn)}

\maketitle

% \vspace{-15pt}
% \begin{flushright}
% \textit{xxxxxxxxxxxxxxxxxxxxxxx \\ xxxxxxxxxxxxxxxxxxxxx \\ -Proverb}
% \end{flushright}

% \vspace{-0.2em}
\begin{figure*}[!htbp]
\centering
\includegraphics[width=1.0\linewidth]{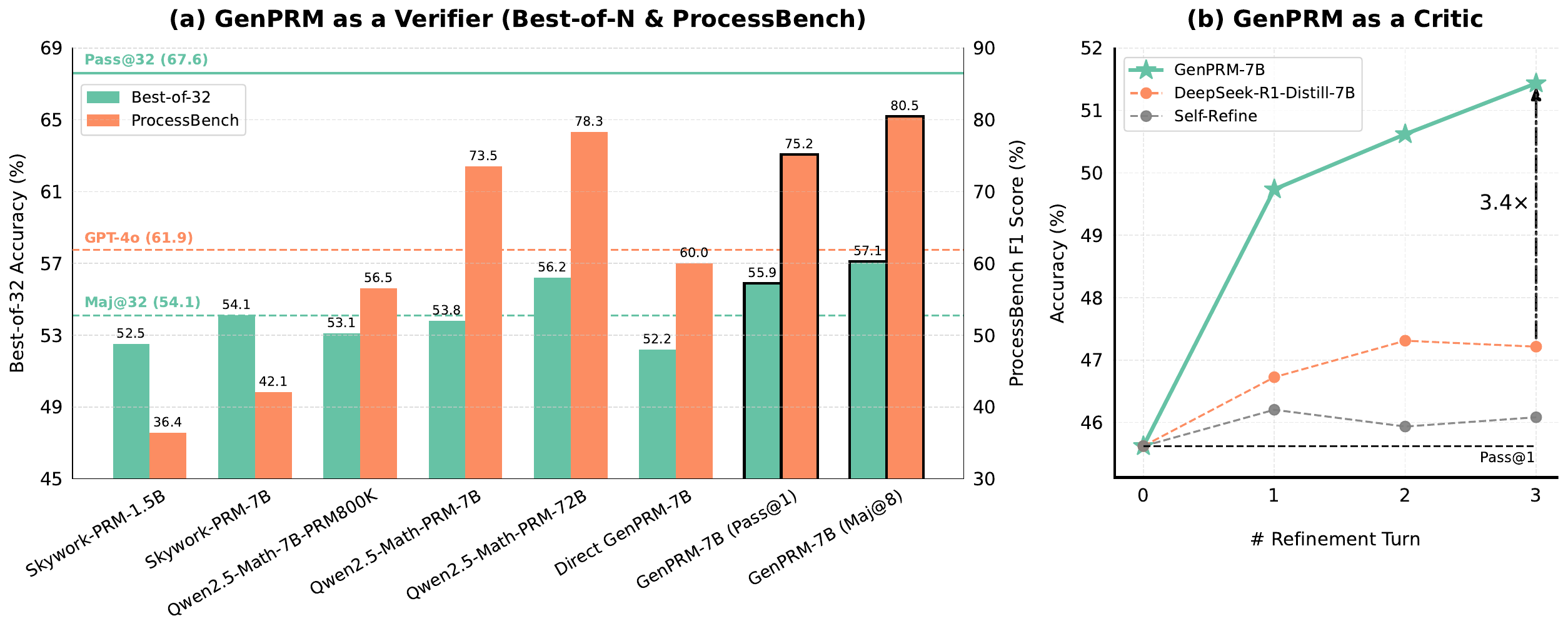}
\vspace{-1.6em}
\caption{\textbf{GenPRM achieves state-of-the-art performance across multiple benchmarks in two key roles:} (a) \textbf{As a verifier}: GenPRM-7B outperforms all classification-based PRMs of comparable size and even surpasses \textbf{Qwen2.5-Math-PRM-72B} via test-time scaling. (b) \textbf{As a critic}: GenPRM-7B demonstrates superior critique capabilities, achieving \textbf{3.4×} greater performance gains than DeepSeek-R1-Distill-Qwen-7B after 3 refinement iterations.}
\vspace{-0.5em}
\label{fig:fig_head}
\end{figure*}

\section{Introduction}

Large Language Models~(LLMs) have shown significant advances in recent years~\citep{GPT-4, Claude, o1, o3, DeepSeek-R1}. As OpenAI o1 demonstrates the great effectiveness of scaling test-time compute~\citep{o1}, an increasing number of researches focus on Test-Time Scaling (TTS) methods to improve the reasoning performance of LLMs~\citep{snell2025scaling, liu2025can}.

Effective TTS requires high-quality verifiers, such as Process Reward Models (PRMs)~\citep{liu2025can}. However, existing PRMs face several limitations. They exhibit limited process supervision capabilities and struggle to generalize across different models and tasks~\citep{ProcessBench, PRMLessons, liu2025can}. Furthermore, most current approaches train PRMs as classifiers that output scalar values, neglecting the natural language generation abilities of LLMs, which are pre-trained on extensive corpora. This classifier-based modeling inherently prevents PRMs from leveraging test-time scaling methods to enhance process supervision capabilities. These limitations lead us to the following research question: \textbf{\textit{How can generative modeling enhance the process supervision capabilities of PRMs while enabling test-time scaling?}}

\begin{figure*}[!t]
\centering
% \vspace{-0.5em}
\includegraphics[width=1.0\textwidth]{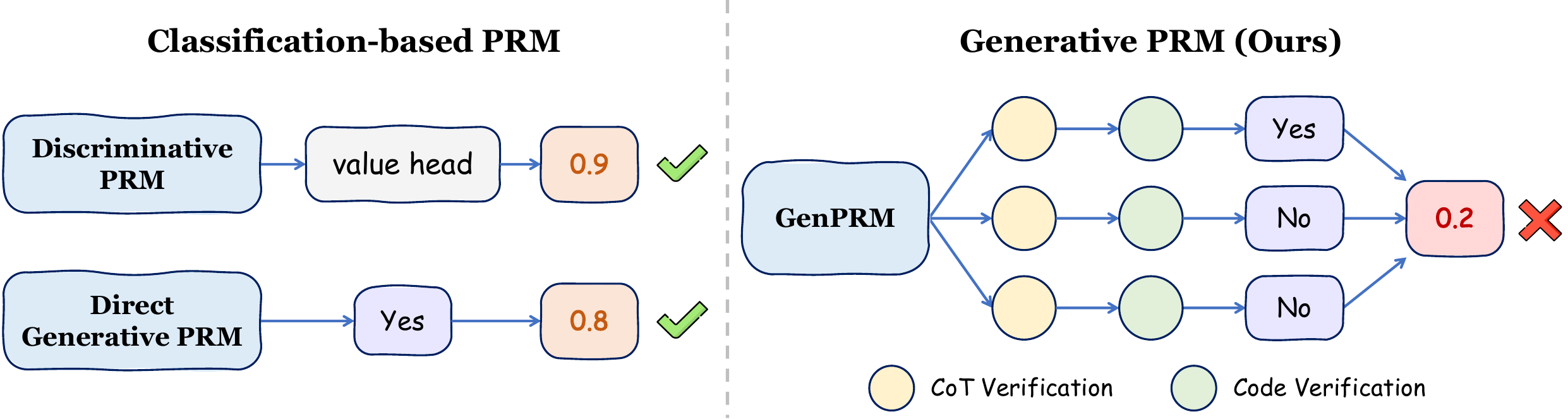}
\caption{Comparison between \ourmethod (right) and previous classification-based PRMs (left).}
% \vspace{-0.5em}
\label{fig:comparison}
\end{figure*}

In this work, we address these challenges through a generative process reward model, named \ourmethod. Specifically, \ourmethod differs from classification-based PRMs in that \ourmethod redefines process supervision as a generative task rather than a discriminative scoring task by integrating Chain-of-Thought~(CoT)~\citep{CoT} reasoning and code verification processes before providing final judgment. To improve conventional hard label estimation, we propose Relative Progress Estimation~(RPE), which leverages a relative criterion for label estimation. Additionally, we introduce a rationale synthesis framework with code verification to obtain high-quality process supervision reasoning data. A comparison of our method with previous classification-based methods is presented in Figure~\ref{fig:comparison}.

Our contributions can be summarized as follows:
\begin{enumerate}
\item We propose a generative process reward model that performs explicit CoT reasoning with code verification and utilizes Relative Progress Estimation to obtain accurate PRM labels.
\item Empirical results on ProcessBench and common mathematical reasoning tasks demonstrate that \ourmethod outperforms prior classification-based PRMs. Additionally, smaller \ourmethod models can surpass larger PRMs via TTS.
\item We provide a new perspective on PRMs in this work, fully leveraging their TTS capabilities, reshaping their applications, and opening new directions for future research in process supervision.
\end{enumerate}

\section{Preliminaries}

\subsection{Markov Decision Process}

Following~\citet{liu2025can}, we formulate the test-time scaling process with PRMs as a Markov Decision Process~(MDP) defined by $(\mathcal{S}, \mathcal{A}, P, r, \gamma)$, where $\mathcal{S}$ is the state space, $\mathcal{A}$ is the action space, $P$ represents transition dynamics, $r: \mathcal{S} \times \mathcal{A} \rightarrow \mathbb{R}$ is the reward function, and $\gamma \in [0, 1]$ is the discount factor. Starting with a prompt set $\mathcal{X}$ and an initial state $s_1 = x \sim \mathcal{X}$, the policy model $\pi_\theta$ generates an action $a_1 \sim \pi_\theta(\cdot \mid s_1)$.\footnote{Following~\citet{snell2025scaling, liu2025can}, we refer to models that generate solutions as policy models.} Unlike traditional RL methods with stochastic transitions~\citep{Meta-Reward-Net, PEARL}, transitions in LLMs are deterministic, i.e., $s_{t+1} = P(\cdot \mid s_t, a_t) = [s_t, a_t]$, where $[\cdot, \cdot]$ denotes string concatenation. This process continues until the episode terminates (i.e., generating the \texttt{[EOS]} token), obtaining a trajectory of $T$ steps: $\tau = \{a_1, a_2, \cdots, a_T\}$. The goal is to optimize either the reward of each step (as in search-based methods) or the reward over the full response (as in Best-of-N sampling).

\subsection{Supervised Fine-Tuning}

Supervised Fine-Tuning~(SFT) trains a model to predict the next token based on prior context. For a dataset $\mathcal{D}_{\text{SFT}} = \{(x^{(i)}, y^{(i)})\}_{i=1}^N$, the SFT loss is:
\begin{equation}
    \mathcal{L}_{\text{SFT}}(\theta) = -\mathbb{E}_{(x, y) \sim \mathcal{D}_{\text{SFT}}} \left[ \sum_{t=1}^{|y|} \log \pi_{\theta}(y_t \mid x, y_{1:t-1}) \right],
\end{equation}
where $\pi_\theta$ represents a model with parameters $\theta$.

\subsection{Test-Time Scaling}

In this work, we consider two test-time scaling methods, including majority voting and Best-of-N.

\paragraph{Majority Voting.}
Majority voting~\citep{Self-Consistency} selects the answer that appears the most frequently among all solutions.

\paragraph{Best-of-N.}
Best-of-N~(BoN)~\citep{brown2024large, snell2025scaling} selects the best answer from $N$ candidate solutions.

\begin{figure*}[!t]
\centering
% \vspace{-0.5em}
\includegraphics[width=1.0\textwidth]{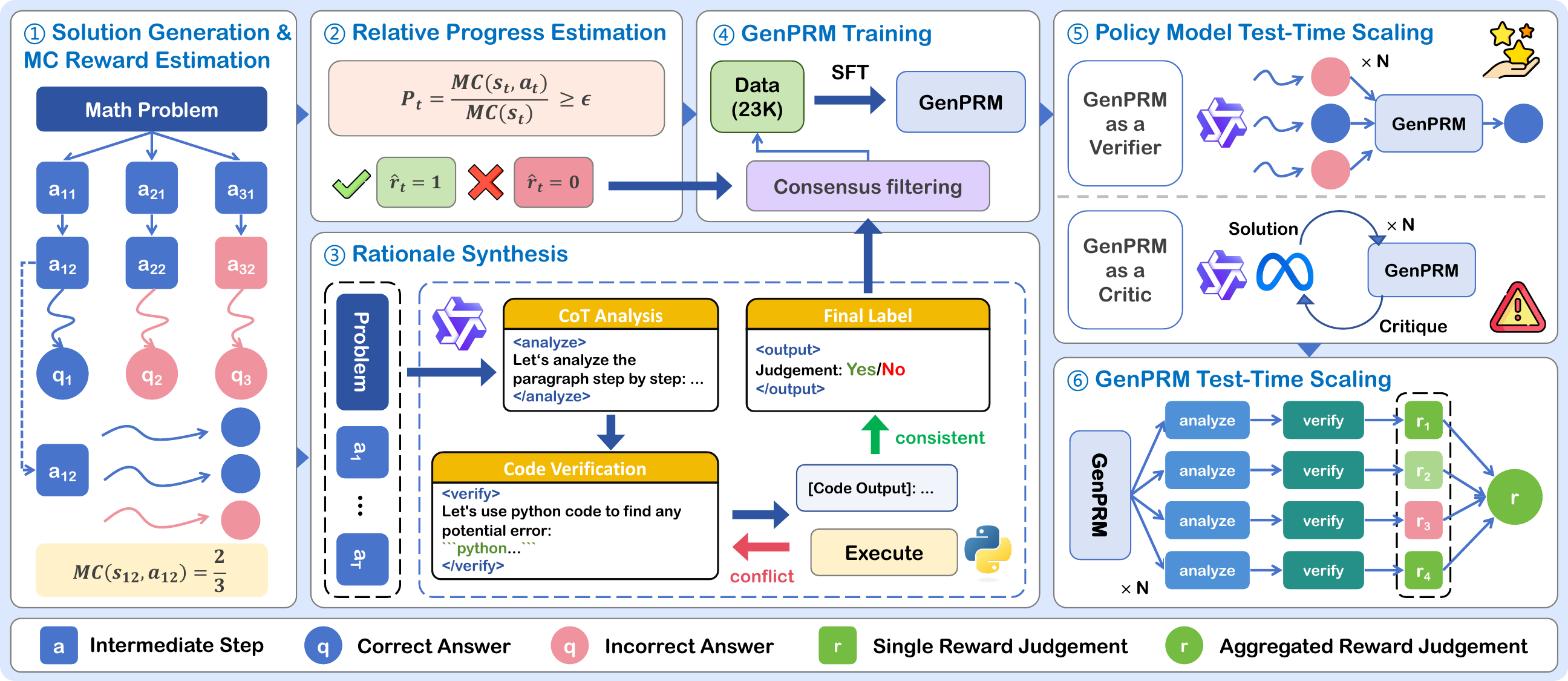}
\caption{\textbf{Overall framework of \ourmethod.} Our framework consists of six key parts:
\textcircled{1} The policy model generates solution steps, with MC scores estimated from rollout trajectories.
\textcircled{2} Our proposed RPE derives accurate PRM labels.
\textcircled{3} High-quality process supervision data is synthesized through CoT reasoning augmented with code verification.
\textcircled{4} We apply consensus filtering followed by SFT to train \ourmethod.
\textcircled{5} The trained \ourmethod functions as a verifier or critic, enabling enhanced test-time scaling for policy models.
\textcircled{6} The performance of \ourmethod further improves through test-time scaling.
}
\vspace{-0.5em}
\label{fig:architecture}
\end{figure*}

\section{Method}
\label{sec:method}

In this section, we first describe how to develop \ourmethod and integrate the reasoning process with code verification. We then introduce how to scale test-time compute of policy models using \ourmethod and apply TTS for \ourmethod. Last, we present the improved label estimation method and data generation and filtering framework of \ourmethod.

\subsection{GenPRM and Test-Time Scaling}
\label{sec:method_genprm}

\subsubsection{From Discriminative PRM to Generative PRM}
\label{ssec:method_disc2gen}

\paragraph{Discriminative PRM.}

Assume we have a PRM dataset $\mathcal{D}_{\text{Disc}} = \{(s_t, a_t), r_t\}$, where $r_t \in \{0, 1\}$ for PRM labels with hard estimation. The discriminative PRM $r_\psi$ is trained via cross-entropy loss~\citep{Skywork-o1-open, PRMLessons}:
\begin{equation}
    \mathcal{L}_{\text{CE}}(\psi) = - \mathbb{E}_{(s_t, a_t, r_t)\sim\mathcal{D}_{\text{Disc}}} \left[ r_t \log r_\psi(s_t, a_t) + (1 - r_t) \log (1-r_\psi(s_t, a_t)) \right].
\end{equation}

\paragraph{Direct Generative PRM.}

With a dataset $\mathcal{D}_{\text{Direct-Gen}} = \left\{(s_t, a_t), r_t \right\}$, where $r_t$ is \texttt{Yes} for a correct step and \texttt{No} otherwise, the direct generative PRM~\citep{RLHFlow} is trained through SFT to predict \texttt{Yes} or \texttt{No} for each step. For step $t$, we use the probability of the \texttt{Yes} token as the predicted process reward $\hat{r}_t$:
\begin{equation}
    \hat{r}_t = r_\psi (\texttt{Yes} \mid s_t, a_t).
\end{equation}

\paragraph{Generative PRM.}

By equipping the direct generative PRM with an explicit reasoning process like CoT~\citep{CoT}, we obtain a generative PRM. Let $v_{1:t-1}$ denote the rationale from step $1$ to $t-1$ and $v_t$ denote the rationale for step $t$. Assume we have a dataset $\mathcal{D}_{\text{Gen}} = \left\{(s_t, a_t, v_{1:t-1}), (v_t, r_t)\right\}$. \ourmethod learns to reason and verify each step via SFT on this dataset. The generative process reward $\hat{r}_t$ can be obtained via the following equation:
\begin{equation}
    \hat{r}_t = r_\psi(\texttt{Yes} \mid s_t, a_t, v_{1:t-1}, v_t), \quad \text{where } v_t \sim r_\psi(\cdot \mid s_t, a_t, v_{1:t-1})
\end{equation}

\paragraph{Generative PRM with Code Verification.}

If we only verify the reasoning step with CoT based on natural language, the process may lack robustness in certain complex scenarios~\citep{PaD, CRITIC}. The difference between the generative PRM and the generative PRM with code verification is that the latter generates code to verify the reasoning step by executing it and provides the judgment based on the execution results. At step $t$, after generating the rationale $v_t$ containing CoT and code, we execute the code and obtain feedback $f_t$. Given the current state $s_t$, action $a_t$, previous rationales $v_{1:t-1}$, and previous corresponding execution feedback $f_{1:t-1}$, the PRM first generates the rationale $v_t$. After execution and obtaining the feedback $f_t$, we compute the final generative process reward as follows:
\begin{equation}
    \hat{r}_t = r_\psi(\texttt{Yes} \mid s_t, a_t, v_{1:t-1}, f_{1:t-1}, v_t, f_t), \quad \text{where } v_t \sim r_\psi(\cdot \mid s_t, a_t, v_{1:t-1}, f_{1:t-1})
\end{equation}
In the following sections, we refer to \ourmethod as this generative PRM type with code verification. The effectiveness of CoT and code verification can be found in Section~\ref{sec:exp_ablation}.

\subsubsection{Test-Time Scaling}
\label{ssec:method_tts}

\paragraph{Policy Model TTS: \ourmethod as a Verifier.}

To scale the test-time compute of policy models, we can sampling multiple responses from policy models and then use \ourmethod as a verifier to select the final answer~\citep{snell2025scaling} in the way of parallel TTS.

\paragraph{Policy Model TTS: \ourmethod as a Critic.}

By equipping the PRM with generative process supervision abilities, \ourmethod can be naturally used as a critic model to refine the outputs of policy models and we can scale the refinement process with multiple turns in a sequential TTS manner.

\paragraph{\ourmethod TTS.}

When evaluating each solution step, we first sample $N$ reasoning verification paths and then use majority voting to obtain the final prediction by averaging the rewards. For \ourmethod without code verification, the rewards are computed as follows:
\begin{equation}
    \hat{r}_t = \frac{1}{N} \sum_{i=1}^N r_\psi(\texttt{Yes} \mid s_t, a_t, v_{1:t-1}^{i}, v_t^{i}).
\end{equation}
And we can further incorporate code verification and execution feedback into this reasoning process:
\begin{equation}
    \hat{r}_t = \frac{1}{N} \sum_{i=1}^N r_\psi(\texttt{Yes} \mid s_t, a_t, v_{1:t-1}^{i}, f_{1:t-1}^{i}, v_t^{i}, f_t^{i}).
\end{equation}
Then the rewards can be used for ranking the responses of policy models or be converted into binary labels through a threshold $0.5$ for judging the correctness of the step. The discussion of using code verification can be found at Table~\ref{tab:abla_component}.

\subsection{Synthesizing Data of GenPRM}
\label{ssec:method_data}

In this section, we introduce our pipeline for synthesizing training data of \ourmethod. The pipeline consists of three stages: (1) generating reasoning paths and obtaining PRM labels via Monte Carlo~(MC) estimation; (2) evaluating the progress of each step via Relative Progress Estimation; and (3) synthesizing rationales with CoT and code verification, and inferring LLM-as-a-judge labels with consensus filtering.

\subsubsection{Solution Generation and Monte Carlo Estimation}
\label{sssec:solution_generation}

\paragraph{Solution Generation with Step Forcing.}
We utilize the 7.5K problems from the training set of the MATH dataset~\citep{MATH} as the problem set. For each problem, we use Qwen2.5-7B-Instruct~\citep{Qwen2.5} as the generation model to collect multiple solutions. Since using ``\verb|\n\n|'' for step division does not consider the semantics of each step and may result in overly fine-grained division, we apply a step forcing approach to generate solutions. Specifically, we add ``\texttt{Step 1:}'' as the prefix for the generation model to complete the response. For a response with $T$ reasoning steps, the format is as follows:
\begin{tcolorbox}[title=The response format with step forcing, label={tab:step_forcing}, breakable, width=\textwidth, fonttitle=\bfseries]
\textbf{Step 1:} \{step content\} \\
... \\
\textbf{Step T:} \{step content\}
\end{tcolorbox}
The proportion of correct paths versus incorrect paths varies significantly depending on the difficulty of the problems. To ensure a sufficient number of correct and incorrect paths, we sample up to $2048$ paths for both hard and easy problems. If no correct or incorrect paths are found after sampling $2048$ responses, we discard the corresponding problems.

\paragraph{Balancing the Precision and Efficiency of MC Estimation.}
Following Math-Shepherd~\citep{Math-Shepherd}, we estimate the probability of correctness for each step using completion-based sampling. For each reasoning step $s_t$, we generate $K$ completion trajectories using a completion model, specifically Qwen2.5-Math-7B-Instruct~\citep{Qwen2.5-Math}, and use MC estimation to calculate the probability that the current step $a_t$ is correct~\citep{Math-Shepherd, PRMLessons}:
\begin{equation}
    MC(s_t, a_t) = MC(s_{t+1}) = \frac{1}{K}\sum_{j=1}^K \mathbbm{1}(q_j = q^*),
\end{equation}
where $q_j$ is the answer of the $j$-th response, $q^*$ is the ground-truth answer, and $\mathbbm{1}$ is the indicator function. However, it is difficult for the completion model to reach the correct answer for hard problems even when the original step is correct, leading to incorrect results for MC estimation. To address this and balance the computation cost, we use a dynamic $K$ based on the estimated Pass@1 $MC(s_1)$:
\begin{equation}
K=
\left\{
\begin{aligned}
    & 128 \quad \ \ \text{if } 0 \le MC(s_1) < 0.1, \\
    & 64 \qquad \text{if } 0.1 \le MC(s_1) < 0.9, \\
    & 32 \qquad \text{if } 0.9 \le MC(s_1) < 1.
\end{aligned}
\right.
\end{equation}

\subsubsection{Relative Progress Estimation}
\label{sssec:RPE}

Previous work has shown that hard label estimation is better than soft label estimation for PRMs~\citep{PRMLessons}. However, after MC estimation, we observe that although the MC score of many steps is greater than 0, the steps are incorrect, as also noted by~\citet{PRMLessons}. On the other hand, we assume that a positive step should be both correct and beneficial. A reasoning step is considered as a beneficial one if it is easier to reach the correct answer by adding this step as the generation prefix. To address these issues, we propose Relative Progress Estimation~(RPE), which shares a similar idea with relative advantage estimation in GRPO~\citep{DeepSeekMath, DeepSeek-R1}, to improve conventional hard label estimation.

Specifically, the MC score is an empirical estimation of the current state $s_t$. To evaluate the quality of the current action $a_t$, it is natural to compare the MC score of the next state $s_{t+1}$ with that of the current state $s_t$, since $s_{t+1} = [s_t, a_t]$. For each response, if the first erroneous step is step $t'$ (i.e., $MC(s_{t'}) = 0$), we set the MC score of the following steps to 0. Our RPE $P_t$ for step $t$ is defined as follows:
\begin{equation}
P_t = \frac{MC(s_t, a_t)}{MC(s_{t})},
\label{eq:RPE}
\end{equation}
where $MC(s_1)$ is the estimated Pass@1 computed in the solution generation phase.
% The relative progress measures 
However, we empirically find that using a strict criterion where progress is always greater than $1$ leads to unsatisfactory performance, as shown in Table~\ref{tab:abla_label}. To address this, we estimate the final reward label $\hat{r}_t$ by introducing a threshold $\epsilon$:
\begin{equation}
\hat{r}_t=
\left\{
\begin{aligned}
    & 1 \qquad \text{if } P_t \ge \epsilon, \\
    & 0 \qquad \text{otherwise.}
\end{aligned}
\right.
\end{equation}
% The relative progress ...
We also discuss another form of relative progress $P_t = MC(s_t, a_t) - MC(s_t)$ in Table~\ref{tab:abla_label} in Section~\ref{sec:exp_ablation}.

\subsubsection{Rationale Generation, Verification and Filtering}
\label{sssec:rationale_generation}

To obtain high-quality rationale data, we use QwQ-32B~\citep{QwQ-32B} as the rationale generation model and introduce a three-step pipeline that automatically generates and verifies the rationale of each reasoning step. Given a problem $x$ with a ground-truth answer $q^*$ and candidate steps $\{a_1, \cdots, a_T\}$, the generation and verification proceed as follows:

\paragraph{Step 1: Code-Based Rationale Generation.}
To evaluate the correctness of $a_t$, we synthesize step-by-step CoT analysis. It has been shown that program-based reasoning improves verification outcomes~\citep{PaD}. Based on CoT analysis, we continue to synthesize code-based rationales to verify $a_t$ based on the problem and historical steps $\{a_1, \cdots, a_{t-1}\}$. We prompt the rationale generation model to surround the CoT with \texttt{<analyze>} and \texttt{</analyze>}, and the code with \texttt{<verify>} and \texttt{</verify>}. The prompt for rationale generation is shown in Table~\ref{tab:rationale_prompt}.

\paragraph{Step 2: Code Execution and Verification.}

With the generated code, we execute it and obtain the feedback $f_t$ for step $t$. The execution feedback is formatted as \texttt{[Code output: \{execution result\}]} and is concatenated to the generated CoT and code as the prefix for the subsequent generation. If the execution result is inconsistent with the generated CoT verification, we observe that QwQ-32B performs self-reflection behaviors until reaching a consensus.

\paragraph{Step 3: Label Judgment and Consensus Filtering.}

After generating and verifying the rationale data of all candidate steps, the rationale generation model finally outputs an number. If all steps are inferred to be correct, the number will be -1, otherwise will be the index of the first erroneous step. For each solution, if there is at least one process label with RPE is not consistent with the labels generated by LLM-as-a-judge~\citep{LLM-as-a-judge}, we discard the entire solution and only retain the one with all labels consistent. After consensus filtering, we discard approximately $51\%$ of the data and finally obtain a dataset containing 23K problems with reasoning steps and rationale data.

\section{Experiments}
\label{sec:experiments}

In this section, we aim to answer the following questions:
\begin{itemize}
    \item \textbf{Q1:} How does \ourmethod perform compared with previous PRMs? (\S\ref{sec:exp_main}, \S\ref{sec:exp_tts})
    \item \textbf{Q2:} How does the performance of \ourmethod scale with more test-time compute? (\S\ref{sec:exp_main}, \S\ref{sec:exp_tts})
    \item \textbf{Q3:} How does \ourmethod benefit policy model test-time scaling? (\S\ref{sec:exp_tts})
    \item \textbf{Q4:} How do the components and hyperparameters influence \ourmethod? (\S\ref{sec:exp_ablation})
\end{itemize}

\subsection{Setup}
\label{sec:exp_setup}

\paragraph{Benchmarks.}

We evaluate \ourmethod and baseline methods on ProcessBench~\citep{ProcessBench}, a benchmark designed to assess process supervision capabilities in mathematical reasoning tasks.\footnote{Our evaluation code is adapted from~\url{https://github.com/QwenLM/ProcessBench}.} Additionally, we conduct BoN and critic refinement experiments using MATH~\citep{MATH}, AMC23~\citep{AMC23}, AIME24~\citep{AIME24}, and Minerva Math~\citep{Minerva-Math}. For BoN response generation, we employ Qwen2.5-Math-7B-Instruct~\citep{Qwen2.5-Math} and Gemma-3-12b-it~\citep{Gemma3} as policy models. For policy model TTS with \ourmethod as the critic, we use Gemma-3-12b-it~\citep{Gemma3} and Qwen2.5-7B-Instruct~\citep{Qwen2.5} as generators.

\paragraph{Baselines.}
For ProcessBench and BoN experiments, we compare \ourmethod with the following methods:
\begin{itemize}
    \item \textbf{Math-Shepherd-PRM-7B}~\citep{Math-Shepherd}: This method trains a PRM using hard labels computed based on MC estimation.
    \item \textbf{RLHFlow series}~\citep{RLHFlow}: Includes RLHFlow-PRM-Mistral-8B and RLHFlow-PRM-Deepseek-8B.
    \item \textbf{Skywork-PRM series}~\citep{Skywork-o1-open}: Comprises Skywork-PRM-1.5B and Skywork-PRM-7B.
    \item \textbf{EurusPRM}~\citep{PRIME}: EurusPRM-Stage1 and EurusPRM-Stage2 are trained as implicit PRMs~\citep{Implicit-PRM}.
    \item \textbf{Qwen2.5-Math series}~\citep{ProcessBench, PRMLessons}: Qwen2.5-Math-7B-Math-Shepherd and Qwen2.5-Math-7B-PRM800K are trained with Math-Shepherd~\citep{Math-Shepherd} and PRM800K~\citep{PRM800K}, respectively. For Qwen2.5-Math-PRM-7B and Qwen2.5-Math-PRM-72B, the training data is applied consensus filtering using LLM-as-a-judge~\citep{LLM-as-a-judge}.  
    \item \textbf{RetrievalPRM-7B}~\citep{RetrievalPRM}: The method enhances PRM with retrieved questions and corresponding steps.
    \item \textbf{Universal-PRM-7B}~\citep{AURORA}: The method proposes an automated framework using ensemble prompting and reverse verification.
    \item \textbf{Dyve-14B}~\citep{Dyve}: This method dynamically applies fast or slow verification for each reasoning step.
    \item \textbf{Direct Generative PRM-7B}: The method trains a direct generative PRM with the original language head via SFT using the same data as \ourmethod, but without CoT and code verification.
\end{itemize}

For critic experiments, we use the following methods for comparison:
\begin{itemize}
    \item \textbf{Self-Refine}~\citep{Self-Refine}: This method uses the generator to self-critique and refine the solution.
    \item \textbf{DeepSeek-R1-Distill-Qwen-7B}~\citep{DeepSeek-R1}: This model is fine-tuned based on Qwen2.5-Math-7B~\citep{Qwen2.5} using high-quality reasoning data generated by DeepSeek-R1~\citep{DeepSeek-R1}.
\end{itemize}

\paragraph{Implementation Details.}

For RPE, we set $\epsilon = 0.8$ across all experiments, with ablation studies presented in Section~\ref{sec:exp_ablation}. Rationale data is generated using QwQ-32B~\citep{QwQ-32B} and the prompt template is shown in Table~\ref{tab:rationale_prompt}. Our base models are from the DeepSeek-R1-Distill series~\citep{DeepSeek-R1}, specifically the 1.5B, 7B, and 32B parameter variants. The training configuration for our method uses a batch size of 64 and a learning rate of $2.0 \times 10^{-6}$. During evaluation, we employ a temperature of 0.6. For critique refinement experiments, we extract content within the \texttt{<analyze></analyze>} tags, focusing exclusively on steps predicted as negative by the policy model. The baseline methods utilize standardized prompt templates (detailed in Table~\ref{tab:critique_prompt}) to ensure consistent critique generation formats.

\subsection{ProcessBench Results}
\label{sec:exp_main}

\paragraph{\ourmethod outperforms classification-based PRMs on ProcessBench.}

As shown in Table~\ref{tab:main_processbench}, \ourmethod-7B significantly outperforms direct generative PRM and surpasses \textbf{all} previous PRMs with parameters less than 72B on ProcessBench. Also, \ourmethod-1.5B outperforms Skywork-PRM-1.5B by a large margin. It is noteworthy that \ourmethod is trained with merely \textbf{23K} data from \textbf{MATH}~\citep{MATH} only. By comparing the detailed results in Table~\ref{tab:all_processbench}, we can find that the performance gain of \ourmethod mainly comes from the stronger abilities of finding \textbf{erroneous} steps and we provide concrete cases in Appendix~\ref{app:cases}. These results demonstrating the superiority of generative modeling of PRM.

\begin{table}[!t]
\centering
\caption{ProcessBench results reported with F1 scores. The results of \ourmethod are \colorbox{\mycolor}{shaded}. For 1.5B PRMs, \textbf{bold} indicates the best Pass@1 or scores superior to GPT-4o. For 7-8B and 14-72B PRMs, \textbf{bold} denotes the best Pass@1 or scores superior to Qwen2.5-Math-PRM-72B.}
\resizebox{0.85\textwidth}{!}{
\begin{tabular}{lcccccc}
\toprule
% \textbf{Model} & \textbf{GSM8K} & \textbf{MATH} & \textbf{OlympiadBench} & \textbf{Omni-MATH} & \textbf{Avg.} \\
\multirow{2}{*}{\textbf{Model}} & \multirow{2}{*}{\textbf{\# Samples}} & \multirow{2}{*}{\textbf{GSM8K}} & \multirow{2}{*}{\textbf{MATH}} & \multirow{2}{*}{\begin{tabular}[c]{@{}c@{}} \bf Olympiad \\ \bf Bench \end{tabular}} & \multirow{2}{*}{\begin{tabular}[c]{@{}c@{}} \bf Omni- \\ \bf MATH \end{tabular}} & \multirow{2}{*}{\textbf{Avg.}} \\
& \\
\midrule
\multicolumn{7}{c}{\textit{Proprietary LLMs (Critic)}} \\
\midrule
GPT-4o-0806                   & unk  & 79.2 & 63.6 & 51.4 & 53.5 & 61.9 \\
o1-mini                       & unk  & 93.2 & 88.9 & 87.2 & 82.4 & 87.9 \\
\midrule
\multicolumn{7}{c}{\textit{PRMs (1.5B)}} \\
\midrule
Skywork-PRM-1.5B              & unk  & \textbf{59.0} & 48.0 & 19.3 & 19.2 & 36.4 \\
\rowcolor{cyan!10} GenPRM-1.5B (Pass@1)          & 23K  & 52.8 & \textbf{66.6} & \textbf{55.1} & \textbf{54.5} & \textbf{57.3} \\
\rowcolor{cyan!10} GenPRM-1.5B (Maj@8)           & 23K  & 51.3 & \textbf{74.4} & \textbf{65.3} & \textbf{62.5} & \textbf{63.4} \\
\midrule
\multicolumn{7}{c}{\textit{PRMs (7-8B)}} \\
\midrule
Math-Shepherd-PRM-7B          & 445K & 47.9 & 29.5 & 24.8 & 23.8 & 31.5 \\
RLHFlow-PRM-Mistral-8B        & 273K & 50.4 & 33.4 & 13.8 & 15.8 & 28.4 \\
RLHFlow-PRM-Deepseek-8B       & 253K & 38.8 & 33.8 & 16.9 & 16.9 & 26.6 \\
Skywork-PRM-7B                & unk  & 70.8 & 53.6 & 22.9 & 21.0 & 42.1 \\
EurusPRM-Stage1               & 463K & 44.3 & 35.6 & 21.7 & 23.1 & 31.2 \\
EurusPRM-Stage2               & 30K  & 47.3 & 35.7 & 21.2 & 20.9 & 31.3 \\
Qwen2.5-Math-7B-Math-Shepherd & 445K & 62.5 & 31.6 & 13.7 & 7.7  & 28.9 \\
Qwen2.5-Math-7B-PRM800K       & 264K & 68.2 & 62.6 & 50.7 & 44.3 & 56.5 \\
Qwen2.5-Math-PRM-7B           & $\sim$344K & 82.4 & 77.6 & 67.5 & 66.3 & 73.5 \\
RetrievalPRM-7B               & 404K & 74.6 & 71.1 & 60.2 & 57.3 & 65.8 \\
Universal-PRM-7B              & unk  & \textbf{85.8} & 77.7 & 67.6 & 66.4 & 74.3 \\
% Discriminative PRM-7B         & 23K  &  &  &  &  &  \\
Direct Generative PRM-7B      & 23K  & 63.9 & 65.8 & 54.5 & 55.9 & 60.0 \\
\rowcolor{cyan!10} GenPRM-7B (Pass@1)            & 23K  & 78.7 & \textbf{80.3} & \textbf{72.2} & \textbf{69.8} & \textbf{75.2} \\
\rowcolor{cyan!10} GenPRM-7B (Maj@8)             & 23K  & 81.0 & \textbf{85.7} & \textbf{78.4} & \textbf{76.8} & \textbf{80.5} \\
\midrule
\multicolumn{7}{c}{\textit{PRMs (14-72B)}} \\
\midrule
% Qwen2.5-Math-RM-72B           & 43.5 & 47.2 & 37.6 & 27.4 & 38.9 \\
Dyve-14B                      & 117K & 68.5 & 58.3 & 49.0 & 47.2 & 55.8 \\
Qwen2.5-Math-PRM-72B          & $\sim$344K & \textbf{87.3} & 80.6 & 74.3 & 71.1 & 78.3 \\
\rowcolor{cyan!10} GenPRM-32B (Pass@1)           & 23K  & 83.1 & \textbf{81.7} & 72.8 & \textbf{72.8} & 77.6 \\
\rowcolor{cyan!10} GenPRM-32B (Maj@8)            & 23K  & 85.1 & \textbf{86.3} & \textbf{78.9} & \textbf{80.1} & \textbf{82.6} \\
\bottomrule
\end{tabular}
}
\label{tab:main_processbench}%
\end{table}%

\paragraph{\ourmethod enables smaller PRMs surpass $10\times$ larger PRMs and GPT-4o via TTS.}
We also compare the TTS results of \ourmethod in Table~\ref{tab:main_processbench} and find that \ourmethod-1.5B surpasses GPT-4 and \ourmethod-7B exceeds Qwen2.5-Math-PRM-72B on ProcessBench via simply majority voting, showing that scaling test-time compute is highly effective for \ourmethod. We also find that the performance improvement of scaling the test-time compute on \textbf{harder} problems is larger than that of easier questions.

\subsection{Policy Model Test-Time Scaling Results}
\label{sec:exp_tts}

\begin{figure*}[!htbp]
% \vspace{-0.5em}
\centering
\begin{tabular}{c}
% \hspace{-2.8em}
\includegraphics[width=0.95\linewidth]{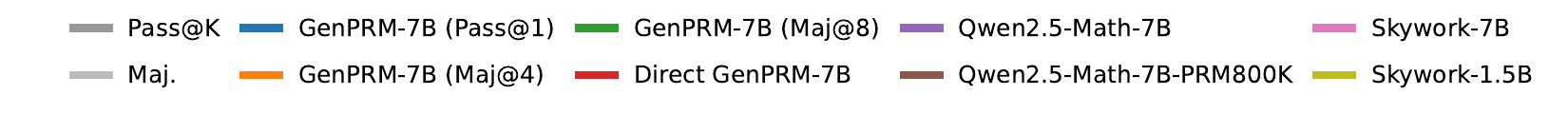}
\vspace{-1.0em}
\end{tabular}

\begin{tabular}{cccc}
\hspace{-1.4em}
\subfloat[\centering MATH (Qwen)]{\centering\includegraphics[width=0.26\linewidth]{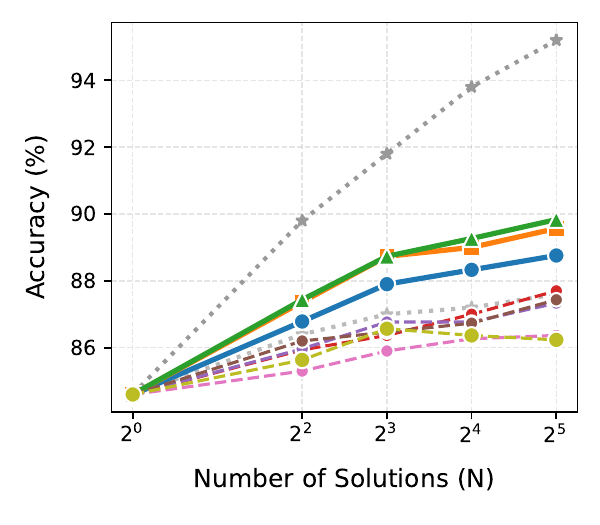}}
\hspace{-1.45em}
& \subfloat[\centering AMC23 (Qwen)]{\includegraphics[width=0.26\linewidth]{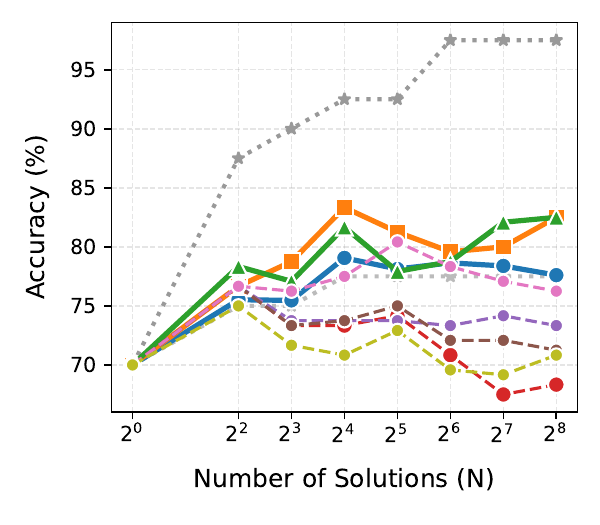}}
\hspace{-1.45em}
& \subfloat[\centering AIME24 (Qwen)]{\includegraphics[width=0.26\linewidth]{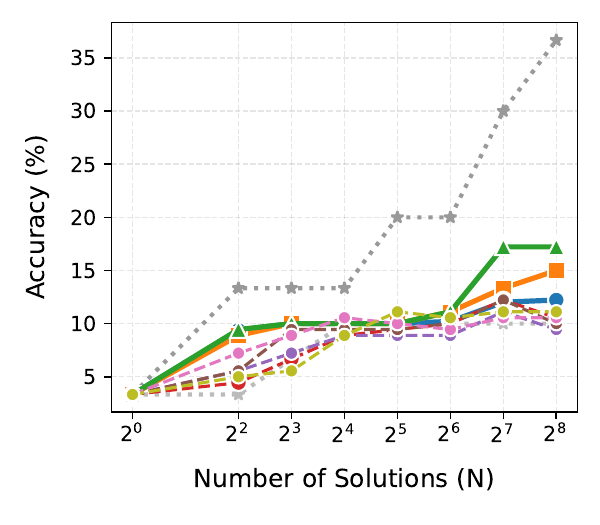}}
\hspace{-1.45em}
& \subfloat[\centering Minerva Math (Qwen)]{\includegraphics[width=0.26\linewidth]{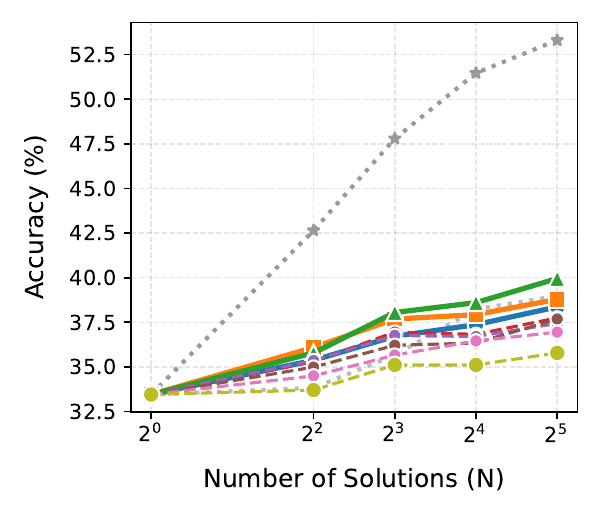}}
% \hspace{-1.45em}
\end{tabular}

% \vspace{0.5em}

\begin{tabular}{cccc}
\hspace{-1.4em}
\subfloat[\centering MATH (Gemma)]{\includegraphics[width=0.26\linewidth]{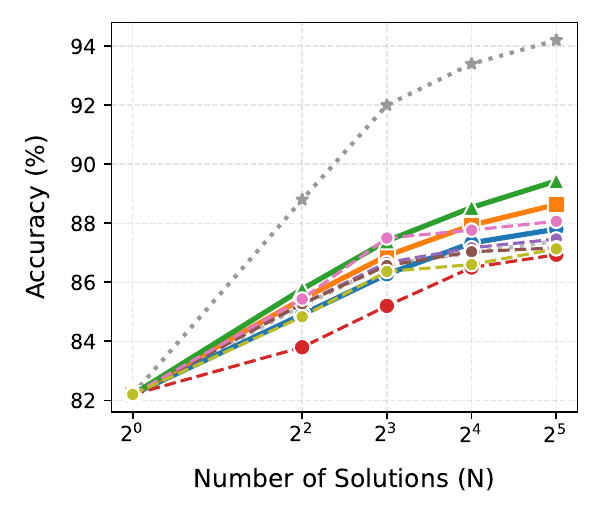}}
\hspace{-1.45em}
& \subfloat[\centering AMC23 (Gemma)]{\includegraphics[width=0.26\linewidth]{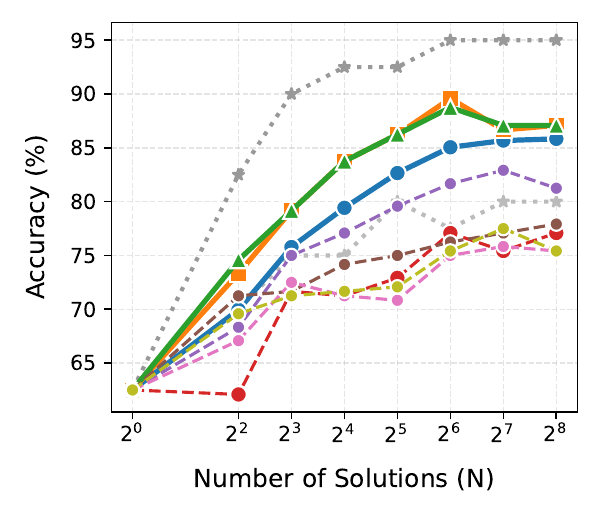}}
\hspace{-1.45em}
& \subfloat[\centering AIME24 (Gemma)]{\includegraphics[width=0.26\linewidth]{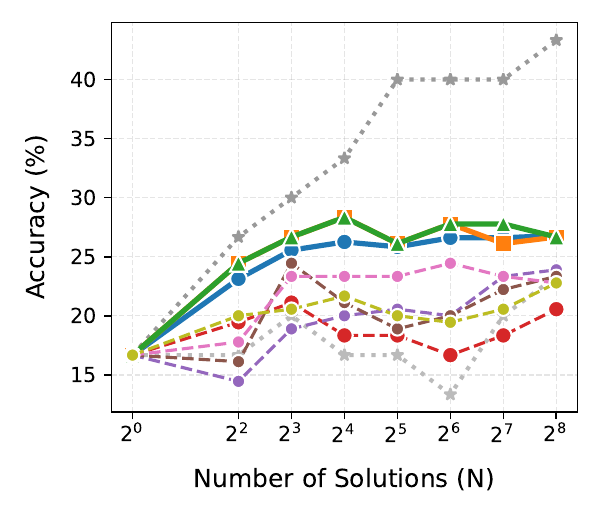}}
\hspace{-1.45em}
& \subfloat[\centering Minerva Math (Gemma)]{\includegraphics[width=0.26\linewidth]{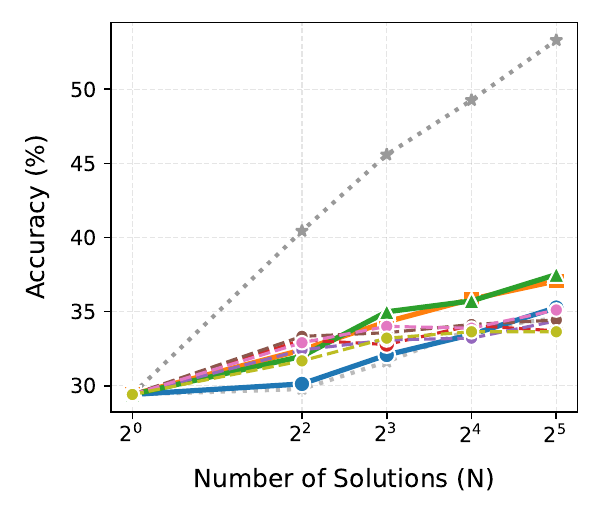}}
% \hspace{-1.45em}
\end{tabular}

\caption{BoN results with different generation models on multiple mathematical benchmarks.}
% \vspace{-0.5em}
\label{fig:main_bon}
\end{figure*}

\paragraph{\ourmethod as a Verifier.}
The results in Figure~\ref{fig:main_bon} (a)-(d) show that \ourmethod outperforms the baselines on MATH, AMC23, AIME24, and Minerva Math with Qwen2.5-Math-7B-Instruct~\citep{Qwen2.5-Math} as the generation model. The advantage of \ourmethod becomes larger by scaling the test-time compute of \ourmethod and the generation model. Figure~\ref{fig:main_bon} (e)-(h) demonstrates that \ourmethod generalizes well to responses with Gemma-3-12b-it~\citep{Gemma3} as the generation model.

\paragraph{\ourmethod as a Critic.}

We also conduct experiments by using \ourmethod as a critic to refine the outputs of the policy model. The results in Table~\ref{tab:critic} and Figure~\ref{fig:fig_head} (right) show that \ourmethod exhibits strong critique abilities than the baselines, significantly improving the performance of the policy model and the performance continues to increase with more refinement based on the critic feedback.

% \begin{table}[!htbp]
% \centering
% \caption{Results of \ourmethod used as a critic model on AMC23.}
% \resizebox{0.4\textwidth}{!}{
% \begin{tabular}{cccc}
% \toprule
% \textbf{Original} & \textbf{Turn 1} & \textbf{Turn 2} & \textbf{Turn 3} \\
% \midrule
% 66.9 & 74.0 & 76.2 & 76.9 \\
% \bottomrule
% \end{tabular}%
% }
% \label{tab:critic}%
% \end{table}%

% \begin{figure*}[!htbp]
% % \vspace{-0.5em}
% \centering
% \begin{tabular}{c}
% % \hspace{-2.8em}
% \end{tabular}
% \begin{tabular}{cc}
% \subfloat[\centering Gemma-12b-it as Generator]{\centering\includegraphics[width=0.45\linewidth]{arXiv/figures/critique_gemma.pdf}}
% % \hspace{4em}
% &\subfloat[\centering Qwen2.5-7B-Instruct as Generator]{\centering\includegraphics[width=0.45\linewidth]{arXiv/figures/critique_qwen.pdf}}
% \end{tabular}
% \caption{\zj{Critique-Refine iteration results with different generation models on average of 4 datasets.}}
% \vspace{-0.5em}
% \label{fig:main_critique}
% \end{figure*}

\begin{table}[!t]
\centering
\caption{Results of critique refinement experiments. The results of \ourmethod are \colorbox{\mycolor}{shaded}. For each refinement turn, the highest values are \textbf{bolded}.}
\resizebox{1.0\textwidth}{!}{
\begin{tabular}{lccccccccccc}
\toprule
\multirow{2}[2]{*}{\textbf{Critic Model}} & \multicolumn{5}{c}{\textit{\textbf{Gemma-3-12b-it as Generator}}} & \multicolumn{5}{c}{\textit{\textbf{Qwen2.5-7B-Instruct as Generator}}} & {\multirow{2}[2]{*}{\large\textbf{Avg.}}} \\
\cmidrule(lr){2-6} \cmidrule(lr){7-11}
& AMC23 & AIME24 & MATH & Minerva Math & \textbf{Avg.} & AMC23 & AIME24 & MATH & Minerva Math & \textbf{Avg.} & \\
\midrule
Zero-shot                    & 64.1 & 15.8 & 83.8 & 31.9 & 48.9 & 51.6 & 7.1  & 76.2 & 34.5 & 42.4 & 45.7 \\
\midrule
\multicolumn{12}{c}{\textit{Turn 1}} \\
\midrule
Generator                    & 66.6 & 15.8 & 84.7 & 33.3 & 50.1 & 50.6 & 8.0  & 76.8 & 34.0 & 42.4 & 46.3 \\
DeepSeek-R1-Distill-7B       & 69.1 & 17.9 & 84.6 & 33.0 & 51.2 & 50.6 & 6.3  & 77.7 & 34.7 & 42.3 & 46.8 \\
\rowcolor{cyan!10} GenPRM-7B & \textbf{74.1} & \textbf{19.6} & \textbf{86.0} & \textbf{35.3} & \textbf{53.8} & \textbf{57.5} & \textbf{8.3}  & \textbf{80.6} & \textbf{36.5} & \textbf{45.7} & \textbf{49.8} \\
\midrule
\multicolumn{12}{c}{\textit{Turn 2}} \\
\midrule
Generator                    & 66.6 & 18.0 & 84.8 & 31.6 & 50.3 & 49.8 & 8.0  & 76.9 & 31.8 & 41.6 & 46.0 \\
DeepSeek-R1-Distill-7B       & 70.9 & 18.3 & 85.0 & 33.5 & 51.9 & 51.9 & 7.9  & 78.1 & 32.8 & 42.7 & 47.3 \\
\rowcolor{cyan!10} GenPRM-7B & \textbf{75.0} & \textbf{21.3} & \textbf{86.9} & \textbf{35.6} & \textbf{54.7} & \textbf{59.4} & \textbf{9.6}  & \textbf{82.2} & \textbf{35.0} & \textbf{46.6} & \textbf{50.7} \\
\midrule
\multicolumn{12}{c}{\textit{Turn 3}} \\
\midrule
Generator                    & 67.8 & 18.1 & 85.0 & 32.1 & 50.8 & 49.7 & 8.1  & 77.1 & 30.8 & 41.4 & 46.1 \\
DeepSeek-R1-Distill-7B       & 69.6 & 18.8 & 85.0 & 33.4 & 51.7 & 51.9 & 8.3  & 78.2 & 32.7 & 42.7 & 47.2 \\
\rowcolor{cyan!10} GenPRM-7B & \textbf{76.2} & \textbf{22.8} & \textbf{86.7} & \textbf{36.0} & \textbf{55.4} & \textbf{62.7} & \textbf{9.3}  & \textbf{82.9} & \textbf{34.9} & \textbf{47.5} & \textbf{51.5} \\
\bottomrule
\end{tabular}%
}
\label{tab:critic}%
\end{table}%

\subsection{Analysis}
\label{sec:exp_ablation}

\paragraph{Label Estimation Method and Criterion.}

To explore how different label estimation influences \ourmethod, we conduct experiments with the following methods: (1) hard label~\citep{Math-Shepherd, PRMLessons}; (2) RPE in~\eqref{eq:RPE}; and (3) a RPE variant ($P_t=MC(s_t,a_t)-MC(s_t)$). For the RPE and its variant, we use different thresholds $\epsilon$ for evaluation and set the labels as correct by checking whether $P_t \ge \epsilon$. The results in Table~\ref{tab:abla_label} show that RPE and its variant outperforms hard label estimation and RPE with $\epsilon=0.8$ achieves the best result. By scaling test-time compute with majority voting, the results in Table~\ref{tab:abla_label_maj8} demonstrate that RPE with $\epsilon=0.8$ still reaches the best.

\begin{table}[!htbp]
\centering
\caption{Results of \ourmethod with different label estimation method and threshold on ProcessBench, reported with Pass@1. The best results are shown in \textbf{bold}.}
\resizebox{0.82\textwidth}{!}{
\begin{tabular}{llccccc}
\toprule
\multirow{2}{*}{\textbf{Estimation Method}} & \multirow{2}{*}{\begin{tabular}[c]{@{}c@{}} \bf Positive Label \\ \bf Criterion \end{tabular}} & \multirow{2}{*}{\textbf{GSM8K}} & \multirow{2}{*}{\textbf{MATH}} & \multirow{2}{*}{\begin{tabular}[c]{@{}c@{}} \bf Olympiad \\ \bf Bench \end{tabular}} & \multirow{2}{*}{\begin{tabular}[c]{@{}c@{}} \bf Omni- \\ \bf MATH \end{tabular}} & \multirow{2}{*}{\textbf{Avg.}} \\
\\
\midrule
$P_t=MC(s_t,a_t)$ (hard label) & $P_t>0$ & 72.9 & 78.9 & \textbf{73.2} & 68.0 & 73.2 \\
\midrule
\multirow{3}{*}{$P_t=MC(s_t,a_t)-MC(s_t)$} & $P_t \ge -0.1$ & 77.3 & 79.9 & 70.8 & 68.5 & 74.1 \\
& $P_t \ge -0.3$ & 76.8 & 79.6 & 71.1 & 69.0 & 74.1 \\
& $P_t \ge -0.5$ & 75.8 & 80.2 & 72.8 & 68.6 & 74.3 \\
\midrule
\multirow{4}{*}{$P_t=\dfrac{MC(s_t,a_t)}{MC(s_t)}$} & $P_t \ge 0.1$ & 74.8 & 78.7 & 71.6 & 68.7 & 73.5 \\
& $P_t \ge 0.5$ & 75.7 & 79.2 & 70.4 & 68.5 & 73.5 \\
& $P_t \ge 0.8$ & \textbf{78.7} & \textbf{80.3} & 72.2 & \textbf{69.8} & \textbf{75.2} \\
& $P_t \ge 1.0$ & 76.4 & 77.4 & 68.1 & 67.2 & 72.3 \\
\bottomrule
\end{tabular}%
}
\label{tab:abla_label}%
\end{table}%

\begin{table}[!htbp]
\centering
\caption{Results of \ourmethod with different label estimation method and threshold on ProcessBench, reported with Maj@8. The best results are shown in \textbf{bold}.}
\resizebox{0.82\textwidth}{!}{
\begin{tabular}{llccccc}
\toprule
\multirow{2}{*}{\textbf{Estimation Method}} & \multirow{2}{*}{\begin{tabular}[c]{@{}c@{}} \bf Positive Label \\ \bf Criterion \end{tabular}} & \multirow{2}{*}{\textbf{GSM8K}} & \multirow{2}{*}{\textbf{MATH}} & \multirow{2}{*}{\begin{tabular}[c]{@{}c@{}} \bf Olympiad \\ \bf Bench \end{tabular}} & \multirow{2}{*}{\begin{tabular}[c]{@{}c@{}} \bf Omni- \\ \bf MATH \end{tabular}} & \multirow{2}{*}{\textbf{Avg.}} \\
\\
\midrule
$P_t=MC(s_t,a_t)$ (hard label) & $P_t>0$ & 75.1 & 83.8 & \textbf{80.6} & 74.4 & 78.5 \\
\midrule
\multirow{3}{*}{$P_t=MC(s_t,a_t)-MC(s_t)$} & $P_t \ge -0.1$ & 79.8 & 85.1 & 78.0 & 74.5 & 79.4 \\
& $P_t \ge -0.3$ & 80.9 & \textbf{86.5} & 78.1 & 75.0 & 80.2 \\
& $P_t \ge -0.5$ & 78.1 & 85.6 & 79.1 & 73.4 & 79.1 \\
\midrule
\multirow{4}{*}{$P_t=\dfrac{MC(s_t,a_t)}{MC(s_t)}$} & $P_t \ge 0.1$ & 77.0 & 84.6 & 78.1 & 75.3 & 78.7 \\
& $P_t \ge 0.5$ & 78.0 & 85.2 & 78.2 & 74.3 & 78.9 \\
& $P_t \ge 0.8$ & 81.0 & 85.7 & 78.4 & \textbf{76.8} & \textbf{80.5} \\
& $P_t \ge 1.0$ & \textbf{81.1} & 84.1 & 76.0 & 74.7 & 79.0 \\
\bottomrule
\end{tabular}%
}
\label{tab:abla_label_maj8}%
\end{table}%

\paragraph{Reasoning Components.}

To understand how each reasoning component influence \ourmethod, we conduct experiments by training \ourmethod with: (1) CoT data only, (2) code verification data only, and (3) full data. During inference phase, we also compare several variants. For example, \ourmethod trained with full data can be used to only verify each step with CoT only by stopping generation at \verb|</analyze>| token. The results in Table~\ref{tab:abla_component} show that: (1) the improvement of \ourmethod mainly comes from CoT reasoning; (2) generating code and reasoning with code execution result improves the process verification performance as well.

\begin{table}[!htbp]
\centering
\caption{Results on ProcessBench of \ourmethod with different reasoning components, reported with Maj@8. The best results are shown in \textbf{bold}.}
\resizebox{0.75\textwidth}{!}{
\begin{tabular}{lllllccccc}
\toprule
\multicolumn{2}{c}{\textbf{Training}} & \multicolumn{3}{c}{\textbf{Inference}} & \multirow{2}[2]{*}{\textbf{GSM8K}} & \multirow{2}[2]{*}{\textbf{MATH}} & \multirow{2}[2]{*}{\begin{tabular}[c]{@{}c@{}} \bf Olympiad \\ \bf Bench \end{tabular}} & \multirow{2}[2]{*}{\begin{tabular}[c]{@{}c@{}} \bf Omni- \\ \bf MATH \end{tabular}} & \multirow{2}[2]{*}{\textbf{Avg.}} \\
\cmidrule(lr){1-2} \cmidrule(lr){3-5}
CoT & Code & CoT & Code & Code Exec. & & & & & \\
\midrule
\ding{55} & \ding{55} & \ding{55} & \ding{55} & \ding{55} & 63.9 & 65.8 & 54.5 & 55.9 & 60.0 \\
\midrule
\ding{55} & \ding{51} & \ding{55} & \ding{51} & \ding{55} & 67.0 & 70.8 & 61.6 & 57.4 & 64.2 \\
\ding{55} & \ding{51} & \ding{55} & \ding{51} & \ding{51} & 70.6 & 76.6 & 67.3 & 63.9 & 69.6 \\
\ding{51} & \ding{55} & \ding{51} & \ding{55} & \ding{55} & 76.4 & 83.0 & \textbf{80.5} & 75.4 & 78.8 \\
\midrule
\multirow{5}{*}{\ding{51}} & \multirow{5}{*}{\ding{51}} & \ding{55} & \ding{51} & \ding{55} & 60.1 & 66.7 & 59.9 & 59.2 & 61.5 \\
& & \ding{55} & \ding{51} & \ding{51} & 61.3 & 74.7 & 68.1 & 62.0 & 66.5 \\
& & \ding{51} & \ding{55} & \ding{55} & 78.8 & 85.1 & 78.7 & 74.9 & 79.3 \\
& & \ding{51} & \ding{51} & \ding{55} & \textbf{81.0} & 85.1 & 78.1 & 75.5 & 79.9 \\
& & \ding{51} & \ding{51} & \ding{51} & \textbf{81.0} & \textbf{85.7} & 78.4 & \textbf{76.8} & \textbf{80.5} \\
\bottomrule
\end{tabular}%
}
\label{tab:abla_component}%
\end{table}%

\section{Related Work}

\paragraph{Process Reward Models.}

Process reward models have been proved to be effective for providing step-wise scores and are superior to outcome reward models in mathematical reasoning tasks~\citep{uesato2022solving, PRM800K}. However, annotating a process supervision dataset such as PRM800K~\citep{PRM800K} requires significant human costs. To mitigate this cost, prior works utilize Monte Carlo estimation~\citep{Math-Shepherd} and binary search~\citep{OmegaPRM} for automated label generation. Subsequent research improves PRMs through methods such as advantage modeling~\citep{PAV}, $Q$-value rankings~\citep{PQM}, implicit entropy regularization~\citep{ER-PRM}, retrieval-augmented generation~\citep{RetrievalPRM}, and fast-slow verification~\citep{Dyve}.
Furthermore, the community has developed high-quality open-source PRMs, including the RLHFlow series~\citep{RLHFlow}, Math-psa~\citep{OpenR}, Skywork series~\citep{Skywork-o1-open}, and Qwen2.5-Math series~\citep{ProcessBench, PRMLessons}. Recently, a line of works focus on extending PRMs to other tasks, including coding~\citep{o1-Coder}, medical tasks~\citep{MedS3}, agentic tasks~\citep{AgentPRM}, general domain tasks~\citep{OpenPRM, VersaPRM}, and multimodal tasks~\citep{VisualPRM}. Current studies also focus on benchmarking PRMs~\citep{ProcessBench, PRMBench} to systematically evaluate their performance.

\paragraph{Large Language Model Test-Time Scaling.}

Scaling test-time computation is an effective method for improving performance during the inference phase~\citep{o1, o3, DeepSeek-R1}. TTS is commonly implemented with external verifiers (e.g., ORMs and PRMs) or strategies (e.g., beam search and MCTS)~\citep{wu2025inference, snell2025scaling, huggingface2024scaling, liu2025can}. In this work, we scale the test-time computation of a generative PRM with an explicit reasoning process and \ourmethod can also serve as a verifier or a critic model in external TTS.

\paragraph{Enhancing the Generative Abilities of Reward Models.}

Previous research has investigated methods to enhance the generative capabilities of reward models using CoT reasoning~\citep{CLoud, GenRM-CoT, GenRM}. For instance, CLoud reward models~\citep{CLoud} are trained to generate critiques for responses and predict rewards using an additional reward head. GenRM-CoT~\citep{GenRM-CoT} and GenRM~\citep{GenRM} train generative reward models that perform CoT reasoning before making final predictions via SFT and preference learning, respectively. CTRL~\citep{CTRL} demonstrates that critic models exhibit strong discriminative abilities when utilized as generative reward models. Prior to these works, GRM~\citep{GRM} regularizes the hidden states of reward models with a text generation loss.

\section{Conclusion}

In this work, we propose \ourmethod, a generative process reward model that performs explicit reasoning and code verification for process supervision and enables scaling the test-time compute of PRMs. Experimental results on ProcessBench and several mathematical datasets show \ourmethod outperforms prior PRMs. We also demonstrate that the performance of \ourmethod increases via test-time scaling and \ourmethod is effective as a critic model. We believe that this work provides perspectives on PRMs by demonstrating the strong TTS abilities of PRMs and extending the applications of PRMs.

\paragraph{Limitations.}

First, \ourmethod provides process supervision by generative reasoning, which introduces additional computation during inference phase. Future work will investigate how to prune the reasoning process dynamically~\citep{Dyve}. Although \ourmethod focuses mainly on mathematical reasoning tasks, it is worth to explore how to apply generative reasoning on coding and general reasoning tasks in the future~\citep{OpenPRM}. Additionally, it would be interesting to leverage RL to incentivize the generative reasoning abilities of \ourmethod.

% \section*{Author Contributions}
% If you'd like to, you may include  a section for author contributions as is done
% in many journals. This is optional and at the discretion of the authors.

% \section*{Acknowledgments}
% Use unnumbered first level headings for the acknowledgments. All
% acknowledgments, including those to funding agencies, go at the end of the paper.

% \section*{Ethics Statement}
% Authors can add an optional ethics statement to the paper. 
% For papers that touch on ethical issues, this section will be evaluated as part of the review process. The ethics statement should come at the end of the paper. It does not count toward the page limit, but should not be more than 1 page. 

% \clearpage
\bibliography{arXiv}

\clearpage
\appendix

\section{Experimental Details}

\subsection{Scoring and Voting Methods}

\paragraph{PRM-Last.}
PRM-Last considers the process reward of the last step of the entire LLM response as the final score, i.e., $\operatorname{score} = r_T$.

\paragraph{PRM-Avg.}
PRM-Avg computes the mean process reward across all steps as the final score, i.e., $\operatorname{score} = \frac{1}{T}\sum_{t=1}^{T} r_t$.

\paragraph{PRM-Min.}
PRM-Min uses the minimum process reward across all steps as the final score, i.e., $\operatorname{score} = \min_r \{r_t\}_{t=1}^T$.

\subsection{Implementation Details}

Prompt for CoT and code rationale generation is shown in Table~\ref{tab:rationale_prompt}.

\begin{tcolorbox}[title=Prompt for CoT and code rationale generation, label={tab:rationale_prompt}, breakable, width=\textwidth,
fonttitle=\bfseries
]
\textbf{[System]:} \\
You are a math teacher. Your task is to review and critique the paragraphs in solution step by step with python code. \\

\textbf{[User]:} \\
The following is the math problem and a solution (split into paragraphs, enclosed with tags and indexed from 1): \\

[Math Problem] \\

\{problem\} \\

[Solution] \\

\textless paragraph\_1\textgreater \\
\{solution\_section\_1\} \\
\textless /paragraph\_1\textgreater \\

... \\

\textless paragraph\_n\textgreater \\
\{solution\_section\_n\} \\
\textless /paragraph\_n\textgreater \\

Your task is to verify the correctness of paragraph in the solution. Split your verification by \verb|`### Paragraph {{ID}}`|. \\

Your verification for each paragraph should be constructed by 2 parts, wrapped by \verb|`<analyze></analyze>`| and \verb|`<verify></verify>`| separately. \\

1. In \verb|`<analyze></analyze>`| part, you need to analyze the reasoning process and explain why the paragraph is correct or incorrect in detail. \\
2. In \verb|`<verify></verify>`| part, you must write **Python code** in the form of \verb|```python\n{{CODE}}\n```| to verify every details that can be verified by code. You can import PyPI (i.e., `sympy`, `scipy` and so on) to implement complicated calculation. Make sure to print the critic results in the code. Every code will be executed automatically by system. You need to analyze the `[Code Output]` after code executing. \\

\textgreater Pay attention that you must follow the format of \verb|```python\n{{CODE}}\n```| when you write the code, otherwise the code will not be executed. \\

After all verifications, if you identify an error in a paragraph, return the **index of the paragraph where the earliest error occurs**. Otherwise, return the **index of -1 (which typically denotes "not found")**. Please put your final answer (i.e., the index) within box in the form of \verb|`$\\boxed{{INDEX}}$`|.
\end{tcolorbox}

% \section{Evaluation Prompt of ProcessBench}

Following~\citet{ProcessBench, PRMLessons}, we use the prompt in Table~\ref{tab:processbench_prompt} to evaluate LLM-as-a-judge methods on ProcessBench~\citep{ProcessBench}. 

\begin{tcolorbox}[title=Evaluation prompt for LLM-as-a-judge methods on ProcessBench, label={tab:processbench_prompt}, breakable, width=\textwidth,
fonttitle=\bfseries
]
% \begin{verbatim}
I will provide a math problem along with a solution. They will be formatted as 
follows: \\

[Math Problem] \\

\textless math\_problem\textgreater \\
...(math problem)... \\
\textless /math\_problem\textgreater \\

[Solution] \\

\textless paragraph\_1\textgreater \\
...(paragraph 1 of solution)... \\
\textless /paragraph\_1\textgreater \\

... \\

\textless paragraph\_n\textgreater \\
...(paragraph n of solution)... \\
\textless /paragraph\_n\textgreater \\

Your task is to review each paragraph of the solution in sequence, analyzing, verifying, and critiquing the reasoning in detail. You need to provide the analyses and the conclusion in the following format: \\

\textless analysis\_1\textgreater \\
...(analysis of paragraph 1)... \\
\textless /analysis\_1\textgreater \\

... \\

\textless analysis\_n\textgreater \\
...(analysis of paragraph n)... \\
\textless /analysis\_n\textgreater \\

\textless conclusion\textgreater \\
Correct/Incorrect \\
\textless /conclusion\textgreater \\

* When you analyze each paragraph, you should use proper verification, recalculation, or reflection to indicate whether it is logically and mathematically valid. Please elaborate on the analysis process carefully. \\

* If an error is detected in any paragraph, you should describe the nature and cause of the error in detail, and suggest how to correct the error or the correct approach. Once a paragraph is found to contain any error, stop further analysis of subsequent paragraphs (as they may depend on the identified error) and directly provide the conclusion of "Incorrect." \\

For instance, given a solution of five paragraphs, if an error is found in the third paragraph, you should reply in the following format: \\

\textless analysis\_1\textgreater \\
...(analysis of paragraph 1)... \\
\textless /analysis\_1\textgreater \\

\textless analysis\_2\textgreater \\
...(analysis of paragraph 2)... \\
\textless /analysis\_2\textgreater \\

\textless analysis\_3\textgreater \\
...(analysis of paragraph 3; since an error is found here, also provide detailed critique and correction guideline)... \\
\textless /analysis\_3\textgreater \\

\textless conclusion\textgreater \\
Incorrect \\
\textless /conclusion\textgreater \\

Note that the analyses of paragraphs 4 and 5 should be skipped as the paragraph 3 has been found to contain an error. \\

* Respond with your analyses and conclusion directly. \\

-------------------------------------------------- \\

The following is the math problem and the solution for you task: \\

[Math Problem] \\

\{tagged\_problem\} \\

[Solution] \\

\{tagged\_response\}
% \end{verbatim}
\end{tcolorbox}

\begin{tcolorbox}[title=Prompt for critique generation, label={tab:critique_prompt}, breakable, width=\textwidth,
fonttitle=\bfseries
]
\textbf{[User]:} \\
The following is a math problem and my solution. Your task is to review and critique the paragraphs in solution step by step. Pay attention that you should not solve the problem and give the final answer. All of your task is to critique. Output your judgement of whether the paragraph is correct in the form of \verb#`\\boxed{{Yes|No}}`# at the end of each paragraph verification:\\

[Math Problem] \\

\{problem\} \\

[Solution] \\

<paragraph\_\{idx\}> \\
\{solution\_section\} \\
</paragraph\_\{idx\}>
\end{tcolorbox}

\section{Additional Results}

We provide full results of ProcessBench in Table~\ref{tab:all_processbench}.

\begin{table}[!t]
\centering
\caption{Full results of critic models and PRMs on ProcessBench.}
\resizebox{\textwidth}{!}{
\begin{tabular}{lccccccccccccc}
% \begin{tabular}{lc@{\hspace{4pt}}c@{\hspace{4pt}}c@{\hspace{4pt}}c@{\hspace{4pt}}c@{\hspace{4pt}}c@{\hspace{4pt}}c@{\hspace{4pt}}c@{\hspace{4pt}}c@{\hspace{8pt}}c@{\hspace{4pt}}c@{\hspace{4pt}}cc}
% \begin{tabular}{lc@{\hspace{4pt}}c@{\hspace{4pt}}cc@{\hspace{4pt}}c@{\hspace{4pt}}cc@{\hspace{7pt}}c@{\hspace{2pt}}cc@{\hspace{4pt}}c@{\hspace{4pt}}cc}
\toprule
\multirow{2}[2]{*}{\textbf{Model}} & \multicolumn{3}{c}{\textbf{GSM8K}} & \multicolumn{3}{c}{\textbf{MATH}} & \multicolumn{3}{c}{\textbf{OlympiadBench}} & \multicolumn{3}{c}{\textbf{Omni-MATH}} & \multirow{2}[2]{*}{\begin{tabular}[c]{@{}c@{}} \bf Avg. \\ \bf F1 \end{tabular}} \\
\cmidrule(lr){2-4} \cmidrule(lr){5-7} \cmidrule(lr){8-10} \cmidrule(lr){11-13}
& Err. & Corr. & \textbf{F1} & Err. & Corr. & \textbf{F1} & Err. & Corr. & \textbf{F1} & Err. & Corr. & \textbf{F1} \\
\midrule
\multicolumn{14}{c}{\textit{Proprietary LLMs (Critic)}} \\
\midrule
GPT-4-0806                    & 70.0 & 91.2  & 79.2 & 54.4 & 76.6 & 63.6 & 45.8 & 58.4 & 51.4 & 45.2 & 65.6 & 53.5 & 61.9 \\
o1-mini                       & 88.9 & 97.9  & 93.2 & 83.5 & 95.1 & 88.9 & 80.2 & 95.6 & 87.2 & 74.8 & 91.7 & 82.4 & 87.9 \\
\midrule
\multicolumn{14}{c}{\textit{Open-Source LLMs (Critic)}} \\
\midrule
Llama-3-8B-Instruct           & 42.5 & 7.8   & 13.1 & 28.6 & 9.1  & 13.8 & 27.1 & 2.7  & 4.8  & 26.1 & 8.3  & 12.6 & 11.1 \\
Llama-3-70B-Instruct          & 35.7 & 96.9  & 52.2 & 13.0 & 93.3 & 22.8 & 12.0 & 92.0 & 21.2 & 11.2 & 91.7 & 20.0 & 29.1 \\
Llama-3.1-8B-Instruct         & 44.4 & 6.2   & 10.9 & 41.9 & 2.7  & 5.1  & 32.4 & 1.5  & 2.8  & 32.0 & 0.8  & 1.6  & 5.1  \\
Llama-3.1-70B-Instruct        & 64.3 & 89.6  & 74.9 & 35.4 & 75.6 & 48.2 & 35.1 & 69.9 & 46.7 & 30.7 & 61.8 & 41.0 & 52.7 \\
Llama-3.3-70B-Instruct        & 72.5 & 96.9  & 82.9 & 43.3 & 94.6 & 59.4 & 31.0 & 94.1 & 46.7 & 28.2 & 90.5 & 43.0 & 58.0 \\
Qwen2.5-Math-7B-Instruct      & 15.5 & 100.0 & 26.8 & 14.8 & 96.8 & 25.7 & 7.7  & 91.7 & 14.2 & 6.9  & 88.0 & 12.7 & 19.9 \\
Qwen2.5-Math-72B-Instruct     & 49.8 & 96.9  & 65.8 & 36.0 & 94.3 & 52.1 & 19.5 & 97.3 & 32.5 & 19.0 & 96.3 & 31.7 & 45.5 \\
Qwen2.5-Coder-7B-Instruct     & 7.7  & 100.0 & 14.3 & 3.4  & 98.3 & 6.5  & 2.1  & 99.1 & 4.1  & 0.9  & 98.3 & 1.8  & 6.7  \\
Qwen2.5-Coder-14B-Instruct    & 33.8 & 96.4  & 50.1 & 25.4 & 92.4 & 39.9 & 20.7 & 94.1 & 34.0 & 15.9 & 94.2 & 27.3 & 37.8 \\
Qwen2.5-Coder-32B-Instruct    & 54.1 & 94.8  & 68.9 & 44.9 & 90.6 & 60.1 & 33.4 & 91.2 & 48.9 & 31.5 & 87.6 & 46.3 & 56.1 \\
Qwen2-7B-Instruct             & 40.6 & 4.7   & 8.4  & 30.5 & 13.8 & 19.0 & 22.4 & 10.9 & 14.7 & 20.0 & 8.7  & 12.1 & 13.6 \\
Qwen2-72B-Instruct            & 57.0 & 82.9  & 67.6 & 37.7 & 70.9 & 49.2 & 34.0 & 55.2 & 42.1 & 32.3 & 53.1 & 40.2 & 49.8 \\
Qwen2.5-7B-Instruct           & 40.6 & 33.2  & 36.5 & 30.8 & 45.1 & 36.6 & 26.5 & 33.9 & 29.7 & 26.2 & 28.6 & 27.4 & 32.6 \\
Qwen2.5-14B-Instruct          & 54.6 & 94.8  & 69.3 & 38.4 & 87.4 & 53.3 & 31.5 & 78.8 & 45.0 & 28.3 & 76.3 & 41.3 & 52.2 \\
Qwen2.5-32B-Instruct          & 49.3 & 97.9  & 65.6 & 36.7 & 95.8 & 53.1 & 25.3 & 95.9 & 40.0 & 24.1 & 92.5 & 38.3 & 49.3 \\
Qwen2.5-72B-Instruct          & 62.8 & 96.9  & 76.2 & 46.3 & 93.1 & 61.8 & 38.7 & 92.6 & 54.6 & 36.6 & 90.9 & 52.2 & 61.2 \\
QwQ-32B-Preview               & 81.6 & 95.3  & 88.0 & 78.1 & 79.3 & 78.7 & 61.4 & 54.6 & 57.8 & 55.7 & 68.0 & 61.3 & 71.5 \\
\midrule
\multicolumn{14}{c}{\textit{PRMs (1.5B)}} \\
\midrule
Skywork-PRM-1.5B              & 50.2 & 71.5  & 59.0 & 37.9 & 65.2 & 48.0 & 15.4 & 26.0 & 19.3 & 13.6 & 32.8 & 19.2 & 36.4 \\
\rowcolor{cyan!10} GenPRM-1.5B (Pass@1)          & 37.0 & 92.7  & 52.8 & 57.1 & 80.1 & 66.6 & 47.0 & 66.5 & 55.1 & 45.2 & 68.7 & 54.5 & 57.3 \\
\rowcolor{cyan!10} GenPRM-1.5B (Maj@8)           & 34.8 & 97.4  & 51.3 & 64.7 & 87.7 & 74.4 & 57.2 & 76.1 & 65.3 & 51.3 & 80.1 & 62.5 & 63.4 \\
\midrule
\multicolumn{14}{c}{\textit{PRMs (7-8B)}} \\
\midrule
Math-Shepherd-PRM-7B          & 32.4 & 91.7  & 47.9 & 18.0 & 82.0 & 29.5 & 15.0 & 71.1 & 24.8 & 14.2 & 73.0 & 23.8 & 31.5 \\
RLHFlow-PRM-Mistral-8B        & 33.8 & 99.0  & 50.4 & 21.7 & 72.2 & 33.4 & 8.2  & 43.1 & 13.8 & 9.6  & 45.2 & 15.8 & 28.4 \\
RLHFlow-PRM-Deepseek-8B       & 24.2 & 98.4  & 38.8 & 21.4 & 80.0 & 33.8 & 10.1 & 51.0 & 16.9 & 10.9 & 51.9 & 16.9 & 26.6 \\
Skywork-PRM-7B                & 61.8 & 82.9  & 70.8 & 43.8 & 62.2 & 53.6 & 17.9 & 31.9 & 22.9 & 14.0 & 41.9 & 21.0 & 42.1 \\
EurusPRM-Stage1               & 46.9 & 42.0  & 44.3 & 33.3 & 38.2 & 35.6 & 23.9 & 19.8 & 21.7 & 21.9 & 24.5 & 23.1 & 31.2 \\
EurusPRM-Stage2               & 51.2 & 44.0  & 47.3 & 36.4 & 35.0 & 35.7 & 25.7 & 18.0 & 21.2 & 23.1 & 19.1 & 20.9 & 31.3 \\
Qwen2.5-Math-7B-Math-Shepherd & 46.4 & 95.9  & 62.5 & 18.9 & 96.6 & 31.6 & 7.4  & 93.8 & 13.7 & 4.0  & 95.0 & 7.7  & 28.9 \\
Qwen2.5-Math-7B-PRM800K       & 53.1 & 95.3  & 68.2 & 48.0 & 90.1 & 62.6 & 35.7 & 87.3 & 50.7 & 29.8 & 86.1 & 44.3 & 56.5 \\
Qwen2.5-Math-PRM-7B           & 72.0 & 96.4  & 82.4 & 68.0 & 90.4 & 77.6 & 55.7 & 85.5 & 67.5 & 55.2 & 83.0 & 66.3 & 73.5 \\
% Qwen2.5-Math-RM-72B           & 41.1 & 46.1  & 43.5 & 39.7 & 58.1 & 47.2 & 28.1 & 56.6 & 37.6 & 18.8 & 50.2 & 27.4 & 38.9 \\
RetrievalPRM-7B               & 64.7 & 88.1  & 74.6 & 67.2 & 75.6 & 71.1 & 56.0 & 65.2 & 60.2 & 52.8 & 62.7 & 57.3 & 65.8 \\
Universal-PRM-7B              &  -   &   -   & 85.8 &  -   &  -   & 77.7 &  -   &  -   & 67.6 &  -   &  -   & 66.4 & 74.3 \\
Direct Generative PRM-7B      & 52.7 & 81.4 & 63.9 & 55.9 & 80.0 & 65.8 & 44.8 & 69.6 & 54.5 & 45.5 & 72.6 & 55.9 & 60.0 \\
\rowcolor{cyan!10} GenPRM-7B (Pass@1)            & 67.7 & 94.0 & 78.7 & 74.6 & 87.0 & 80.3 & 68.3 & 76.6 & 72.2 & 63.5 & 77.4 & 69.8 & 75.2 \\
\rowcolor{cyan!10} GenPRM-7B (Maj@8)             & 69.6 & 96.9 & 81.0 & 80.5 & 91.6 & 85.7 & 74.0 & 83.5 & 78.4 & 70.0 & 85.1 & 76.8 & 80.5 \\
\midrule
\multicolumn{14}{c}{\textit{PRMs (14-72B)}} \\
\midrule
Dyve-14B                      &  -   &   -   & 68.5 &  -   &  -   & 58.3 &  -   &  -   & 49.0 &  -   &  -   & 47.2 & 55.8 \\
Qwen2.5-Math-PRM-72B          & 78.7 & 97.9  & 87.3 & 74.2 & 88.2 & 80.6 & 67.9 & 82.0 & 74.3 & 64.8 & 78.8 & 71.1 & 78.3 \\
\rowcolor{cyan!10} GenPRM-32B (Pass@1)           & 73.1 & 96.4 & 83.1 & 79.4 & 84.1 & 81.7 & 73.4 & 72.2 & 72.8 & 70.3 & 75.5 & 72.8 & 77.6 \\
\rowcolor{cyan!10} GenPRM-32B (Maj@8)            & 74.9 & 98.5 & 85.1 & 84.0 & 88.7 & 86.3 & 79.0 & 78.8 & 78.9 & 76.3 & 84.2 & 80.1 & 82.6 \\
\bottomrule
\end{tabular}
}
\label{tab:all_processbench}
\end{table}

\paragraph{Model Size.}

We investigate the impact of model size on \ourmethod by evaluating variants with 1.5B, 7B, and 32B parameters. As shown in Table~\ref{tab:abla_model_size}, scaling the model from 1.5B to 7B parameters yields substantial performance gains (57.3 $\rightarrow$ 75.2 and 63.4 $\rightarrow$ 80.5). However, further increasing the model size to 32B provides only marginal improvements, suggesting that the 7B variant offers the best balance between efficiency and effectiveness.

\begin{table}[!htbp]
\centering
\caption{Evaluation results of \ourmethod with different sizes on ProcessBench.}
\resizebox{0.8\textwidth}{!}{
\begin{tabular}{llccccc}
\toprule
\textbf{Model Size} & \textbf{Metric} & \textbf{GSM8K} & \textbf{MATH} & \textbf{OlympiadBench} & \textbf{Omni-MATH} & \textbf{Avg.} \\
\midrule
\multirow{2}{*}{1.5B} & Pass@1 & 52.8 & 66.6 & 55.1 & 54.5 & 57.3 \\
                      & Maj@8  & 81.0 & 74.4 & 65.3 & 62.5 & 63.4 \\
\multirow{2}{*}{7B}   & Pass@1 & 78.7 & 80.3 & 72.2 & 69.8 & 75.2 \\
                      & Maj@8  & 81.0 & 85.7 & 78.4 & 76.8 & 80.5 \\
\multirow{2}{*}{32B}  & Pass@1 & 83.1 & 81.7 & 72.8 & 72.8 & 77.6 \\
                      & Maj@8  & 85.1 & 86.3 & 78.9 & 80.1 & 82.6 \\
\bottomrule
\end{tabular}
}
\label{tab:abla_model_size}%
\end{table}%

\paragraph{Data Size.}

To assess the influence of training data volume, we train \ourmethod on progressively larger subsets of ProcessBench (25\%, 50\%, and 100\% of the full dataset). Table~\ref{tab:abla_data_size} demonstrates that Pass@1 F1 scores improve rapidly with initial data increases, but the growth rate slows substantially with additional data.

\begin{table}[!htbp]
\centering
\caption{Evaluation results of GenPRM with different proportions of training data on ProcessBench.}
\resizebox{0.8\textwidth}{!}{
\begin{tabular}{lcccccc}
\toprule
\textbf{Data Size} & \textbf{Metric} & \textbf{GSM8K} & \textbf{MATH} & \textbf{OlympiadBench} & \textbf{Omni-MATH} & \textbf{Avg.} \\
\midrule
25\%  & Pass@1 & 71.8  & 78.0  & 68.9  & 68.2  & 71.7  \\
50\%  & Pass@1 & 78.1  & 80.6  & 71.5  & 67.6  & 74.5  \\
100\% & Pass@1 & 78.7  & 80.3  & 72.2  & 69.8  & 75.2  \\
\bottomrule
\end{tabular}%
}
\label{tab:abla_data_size}%
\end{table}%

\paragraph{Inference Tokens.}

We provide statistics of the reasoning tokens per step and per response in Table~\ref{tab:token}.

\begin{table}[!htbp]
\centering
\caption{Statistics of the output tokens of \ourmethod.}
\resizebox{0.55\textwidth}{!}{
\begin{tabular}{lcccc}
\toprule
& \textbf{MATH} & \textbf{AMC23} & \textbf{AIME24} & \textbf{Minerva Math} \\
\midrule
Step & 344.7 & 416.2 & 432.5 & 503.3 \\
Response & 2771.4 & 3200.2 & 4112.9 & 4877.1 \\
\bottomrule
\end{tabular}%
}
\label{tab:token}%
\end{table}%

\section{Cases}
\label{app:cases}

\begin{figure*}[!t]
\centering
\vspace{-1em}
\includegraphics[width=0.7\textwidth]{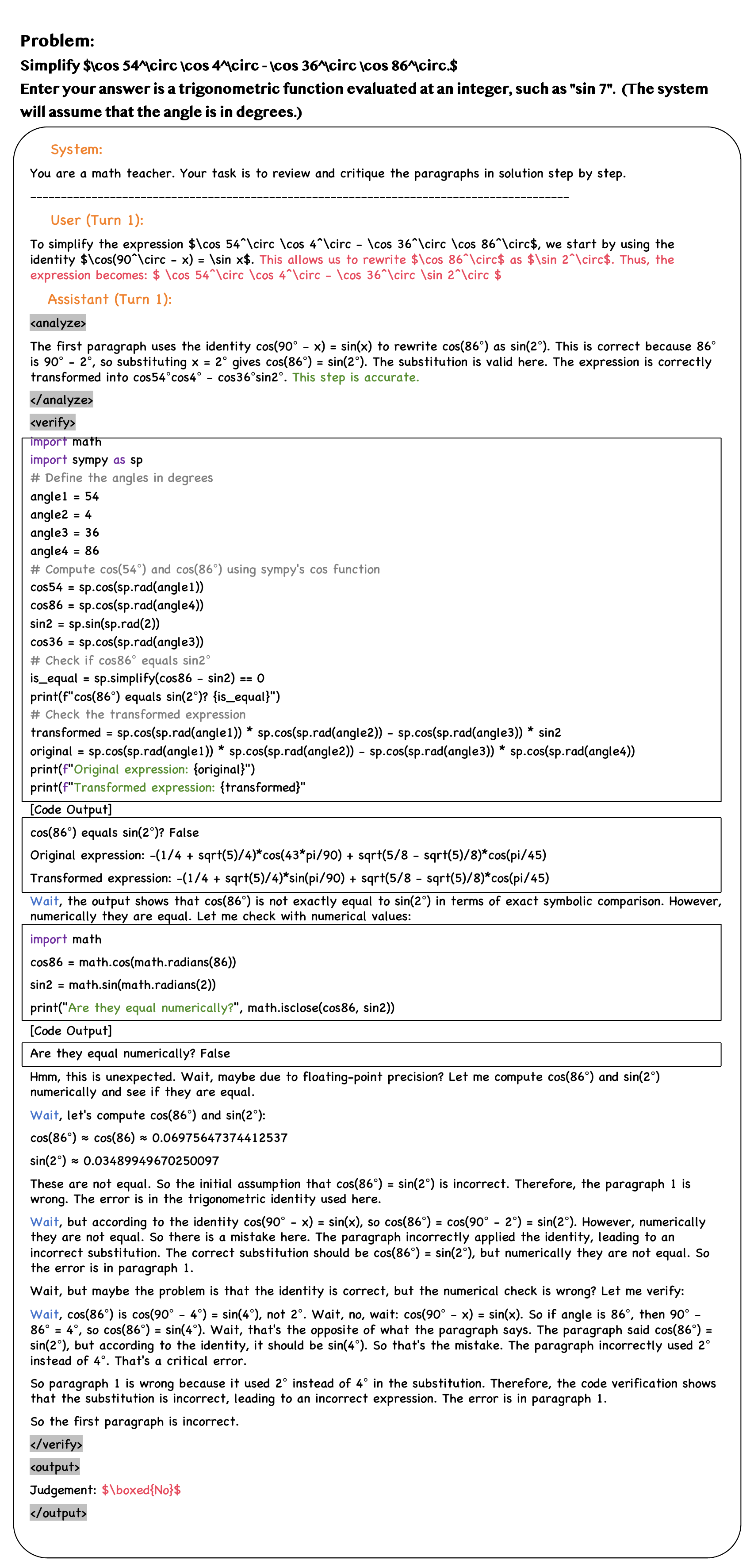}
\caption{The first case of \ourmethod.}
\vspace{-1em}
\label{fig:case1}
\end{figure*}

\begin{figure*}[!t]
\centering
\vspace{-1em}
\includegraphics[width=0.9\textwidth]{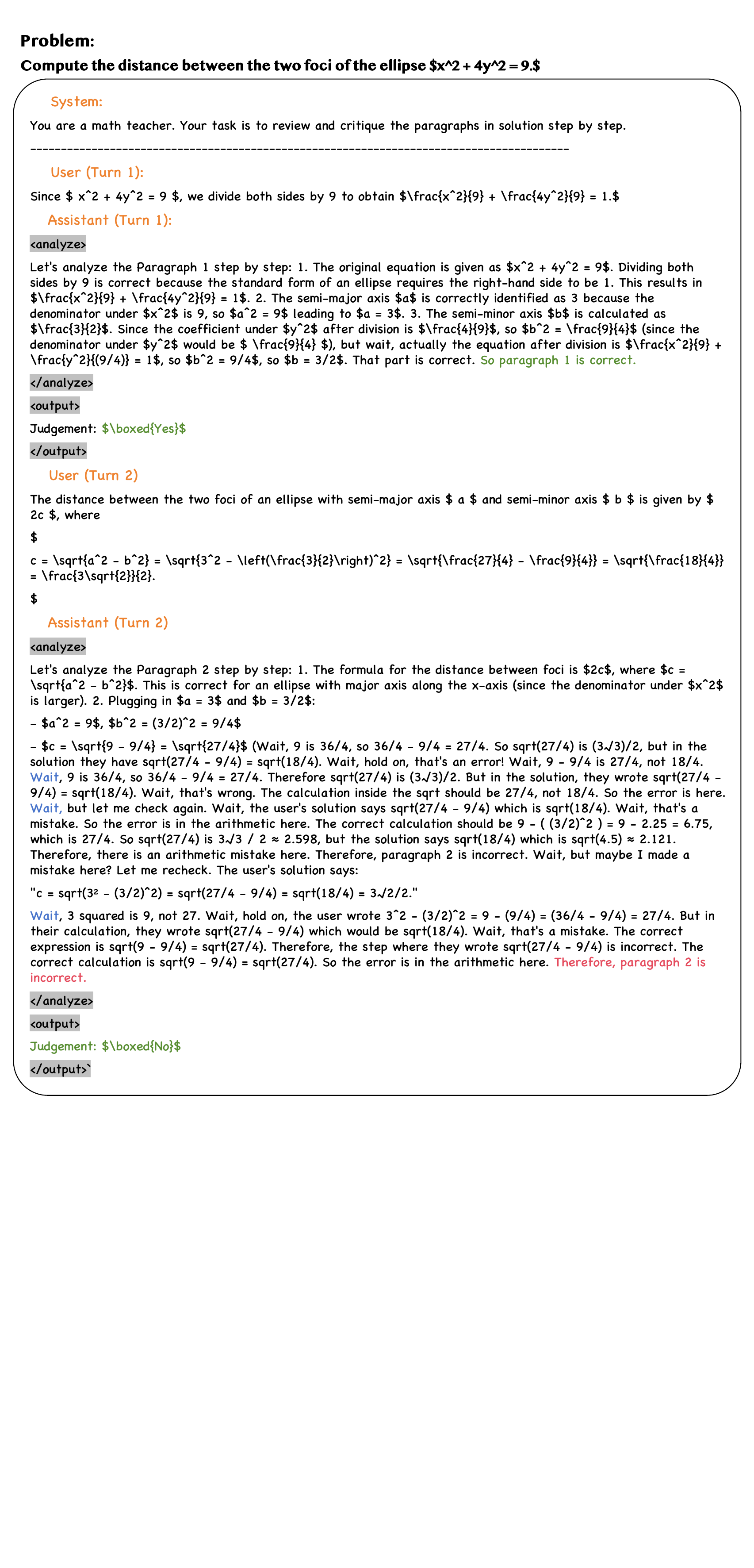}
\vspace{-24em}
\caption{The second case of \ourmethod.}
\vspace{-1em}
\label{fig:case2}
\end{figure*}

In this section, we analyze two cases to have a better understanding of \ourmethod. The case in Figure~\ref{fig:case1} shows that the code execution feedback can correct the mistakes in CoT and enhance the process supervision abilities of \ourmethod. The case in Figure~\ref{fig:case2} demonstrates that \ourmethod provides accurate process supervision with CoT reasoning only.

% \suppressfloats[tbp]

% \newlength{\pdfheight}
% \begin{figure*}[!htbp]
% \centering
% \vspace{-1em}
% \settowidth{\pdfheight}{\includegraphics{figures/case1.pdf}}
% \includegraphics[
%     width=0.9\textwidth,
%     trim=0 1.5\pdfheight{} 0 0,  % 自动计算中点
%     clip
% ]{figures/case1.pdf}
% % \caption{The first case of \ourmethod (upper half).}
% \vspace{-1em}
% \label{fig:case1}
% \end{figure*}
 
% \newlength{\pdfheight}
% \afterpage{
% \begin{figure*}[!htbp]
% \centering
% \vspace{-1em}
% \settowidth{\pdfheight}{\includegraphics{figures/case1.pdf}}
% \includegraphics[
%     width=0.9\textwidth,
%     trim=0 0 0 0.57\pdfheight{},  % 自动计算中点
%     clip
% ]{figures/case1.pdf}
% \caption{The first case of \ourmethod}
% \vspace{-1em}
% % \label{fig:case1}
% \end{figure*}
% }

\end{document}